# „Es geht um Respekt, nicht um Technologie": Erkenntnisse aus einem Interessensgruppen-übergreifenden Workshop zu genderfairer Sprache und Sprachtechnologie

Sabrina Burtscher
Katta Spiel
Human Computer Interaction Group,
TU Wien
Wien, Österreich

Lukas Daniel Klausner
Institut für IT-Sicherheitsforschung
sowie Center for Artificial
Intelligence, FH St. Pölten
St. Pölten, Österreich

Manuel Lardelli
Dagmar Gromann
Zentrum für
Translationswissenschaft, Universität
Wien
Wien, Österreich

## ZUSAMMENFASSUNG

**English:** With the increasing attention non-binary people receive in Western societies, strategies of gender-fair language have started to move away from binary (only female/male) concepts of gender. Nevertheless, hardly any approaches to take these identities into account into machine translation models exist so far. A lack of understanding of the socio-technical implications of such technologies risks further reproducing linguistic mechanisms of oppression and mislabelling. In this paper, we describe the methods and results of a workshop on gender-fair language and language technologies, which was led and organised by ten researchers from TU Wien, St. Pölten UAS, FH Campus Wien and the University of Vienna and took place in Vienna in autumn 2021. A wide range of interest groups and their representatives were invited to ensure that the topic could be dealt with holistically. Accordingly, we aimed to include translators, machine translation experts and non-binary individuals (as "community experts") on an equal footing. Our analysis shows that gender in machine translation requires a high degree of context sensitivity, that developers of such technologies need to position themselves cautiously in a process still under social negotiation, and that flexible approaches seem most adequate at present. We then illustrate steps that follow from our results for the field of gender-fair language technologies so that technological developments can adequately line up with social advancements.

**Deutsch:** Mit zunehmender gesamtgesellschaftlicher Wahrnehmung nicht-binärer Personen haben sich in den letzten Jahren auch Konzepte von genderfairer Sprache von der bisher verwendeten Binarität (weiblich/männlich) entfernt. Trotzdem gibt es bislang nur wenige Ansätze dazu, diese Identitäten in maschineller Übersetzung abzubilden. Ein fehlendes Verständnis unterschiedlicher sozio-technischer Implikationen derartiger Technologien birgt in sich die Gefahr, fehlerhafte Ansprachen und Bezeichnungen sowie sprachliche Unterdrückungsmechanismen zu reproduzieren. In diesem Beitrag beschreiben wir die Methoden und Ergebnisse eines Workshops zu genderfairer Sprache in technologischen Zusammenhängen, der im Herbst 2021 in Wien stattgefunden hat. Zehn Forscher*innen der TU Wien, FH St. Pölten, FH Campus Wien und Universität Wien organisierten und leiteten den Workshop. Dabei wurden unterschiedlichste Interessensgruppen und deren Vertreter*innen breit gestreut eingeladen, um sicherzustellen, dass das Thema holistisch behandelt werden kann. Dementsprechend setzten wir uns zum Ziel, Machine-Translation-Entwickler*innen, Übersetzer*innen, und nicht-binäre Privatpersonen (als „Lebenswelt-Expert*innen") gleichberechtigt einzubinden. Unsere Analyse zeigt, dass Geschlecht in maschineller Übersetzung eine maßgeblich kontextsensible Herangehensweise erfordert, die Entwicklung von Sprachtechnologien sich vorsichtig in einem sich noch in Aushandlung befindlichen gesellschaftlichen Prozess positionieren muss, und flexible Ansätze derzeit am adäquatesten erscheinen. Wir zeigen auf, welche nächsten Schritte im Bereich genderfairer Technologien notwendig sind, damit technische mit sozialen Entwicklungen mithalten können.

# 1 EINLEITUNG

In den letzten Jahren nimmt die gesellschaftliche Wahrnehmung von nicht-binären, intergeschlechtlichen und genderqueeren[1] Personen kontinuierlich zu. Dies zeigt sich nicht zuletzt in den aktualisierten gesetzlichen Rahmenbedingungen, beispielsweise in Österreich und Deutschland, wo seit 2018/2019 ein legaler nicht-binärer Geschlechtseintrag (*divers* bzw. *divers*/*inter*/*offen*/*kein Eintrag*) in Geburtsurkunden und Pässen möglich ist [13, 20]. Ebenso nehmen beispielsweise im Rahmen der Populärkultur nicht-binäre Personen auch im öffentlichen Leben mehr und mehr Raum ein, nicht zuletzt, weil ein gesamtgesellschaftliches Verständnis dazu langsam mehr Freiraum zulässt [33]. Damit einhergehend werden auf sprachlicher Ebene verschiedene Strategien zur Überwindung von Gender Bias im Deutschen jenseits eines binären Geschlechterkonzepts (männlich/weiblich: z. B. LeserInnen, Leser(innen)) vorgeschlagen, wie etwa *geschlechtsinklusive* (z. B. Leser*innen) oder *geschlechtsneutrale* Sprache (z. B. Lesende); wir fassen die letzten beiden Strategien im Folgenden unter der Bezeichnung *genderfaire Sprache* zusammen [36]. Im Zuge der grundlegenden rechtlichen Anerkennung wurden auch offizielle Empfehlungen und Leitlinien für gendersensible Sprache entwickelt, beispielsweise seitens des Rats für deutsche Rechtschreibung im Jahre 2018 als kurze Empfehlungsliste [34] oder seitens der österreichischen Gleichbehandlungsanwaltschaft im Juni 2021 in Form eines Leitfadens zu geschlechtersensibler Sprache [17].[2] Diese dienen auch als praktische Handlungsanleitung für eine inklusive und respektvolle Kommunikation, unabhängig von eigener Identität, körperlichen Geschlechtsmerkmalen und Geschlechtsausdruck, gehen dabei aber maßgeblich auf *geschriebene* und *natürliche* Sprache ein. Derartige gesellschaftliche Entwicklungen werfen Fragen für die Gestaltung und Verwendung von Sprachtechnologien einschließlich der maschinellen Übersetzung (MT; von Englisch Machine Translation) auf. In Bereichen der Sprachdienstleistung und Sprachtechnologie beschränkt sich die Diskussion zu genderfairer Sprache bisher jedoch meist noch auf eine binäre Vorstellung von Geschlecht im Sinne einer ausschließlichen Berücksichtigung von männlich und weiblich [35]. Um Sprachtechnologien demnach auf die Höhe gegenwartsgesellschaftlicher Sprachverhandlungen zu bringen, braucht es holistische Herangehensweisen, die die Perspektiven von Minderheitengeschlechtern, Sprachexpert*innen[3] sowie technischen Entwickler*innen miteinander in Verbindung setzen.

Um das Thema näher zu erörtern, organisierten Forscher*innen der TU Wien, FH St. Pölten, FH Campus Wien und Universität Wien im Herbst 2021 einen dreitägigen Workshop zur partizipativen Erforschung der unterschiedlichen ineinander greifenden Aspekte der gesellschaftlichen, individuellen, technischen und sprachlichen Komponenten sowie eine daran anschließende öffentliche Podiumsdiskussion mit einer Reihe an diversen Interessensvertreter*innen. An beiden Veranstaltungen nahmen Personen der nicht-binären und queeren[4] Gemeinschaften,[5] professionelle Übersetzer*innen sowie Entwickler*innen und Forscher*innen im Bereich maschineller Übersetzungssysteme teil.

Das Ziel des Workshops war ein Gedankenaustausch zwischen Vertreter*innen dieser drei Gruppen, um einen ersten Schritt in Richtung eines Verständnisses für eine Anforderungsanalyse von genderfairer maschineller Übersetzung zu setzen. In interaktiven Einheiten wurden gemeinsam über drei Tage hinweg Strategien für genderfaire Sprache und damit einhergehende Sprachtechnologien mit besonderem Augenmerk auf maschineller Übersetzung erarbeitet. Diskussionen gingen dabei weit über den Rahmen der genderfairen maschinellen Übersetzung (GenderFairMT, so auch der Kurztitel des Workshops) hinaus und befassten sich auch mit allgemeineren Überlegungen zum Thema genderfaire Sprache im deutschen Sprachraum, was wiederum aufzeigt, dass sich die gesamtgesellschaftlichen Aushandlungsprozesse auch in der Gestaltung von technologischen Ansätzen widerspiegeln.

Wir beginnen mit einer kurzen Einführung in linguistische und technische Grundlagen von Humanübersetzung und maschineller Übersetzung, präsentieren unsere Herangehensweise an die Gestaltung des partizipativen Workshops, die Hauptthemen und Ergebnisse aus dem Workshop sowie die wichtigsten Diskussionspunkte der Podiumsdiskussion. Zuletzt bieten wir eine Detailanalyse und Schlussfolgerungen sowie einen Ausblick auf weitere Aktivitäten. In diesem Dokument verwenden wir dabei absichtlich in jedem Abschnitt unterschiedliche Strategien neutraler und inklusiver Sprache, um der Vielfalt relevanter Ansätze aus unserer Arbeit gerecht zu werden. Kurze Informationen zu den verschiedenen Strategien finden sich dabei in den zugehörigen Fußnoten.

# 2 HINTERGRUND

Unsere Arbeit baut auf Erkenntnissen auf, die darlegen, wie in der deutschen Sprache Geschlecht bisher verhandelt wurde. Zudem ziehen wir einige Strategien für genderfaire Sprache heran, die aktuell eine gewisse Verbreitung und Bekanntheit haben. In diesem Kontext betrachten wir außerdem einige Auswirkungen von Stereotypen reproduzierender Sprache auf Menschen und maschinelle Übersetzung.

## 2.1 Geschlecht in der deutschen Sprache

Natürliche Sprachen werden eingeteilt in Sprachen mit (1) **grammatischem Geschlecht**, (2) **angenommenem Geschlecht**[6] und (3) Sprachen **ohne Geschlecht** [35]. Deutsch gilt als Sprache mit grammatischem Geschlecht (ebenso wie beispielsweise Italienisch oder Französisch). In solchen Sprachen wird jedem Nomen ein Geschlecht zugeordnet. Damit werden sowohl Lebewesen als auch Objekte und abstrakte Konzepte mit einem Geschlecht versehen. Wortarten abseits von Nomen, wie Artikel und Adjektive, werden entsprechend gebeugt. Im Unterschied dazu verwenden Sprachen mit angenommenem Geschlecht (z. B. Dänisch oder Englisch) gegenderte Ausdrücke ausschließlich in der Form von lexikalischem Geschlecht (z. B. *girl/boy*), bei manchen zusammengesetzten Wörtern (z. B. *chairman/chairwoman*) und Pronomina (z. B. *she/he/it*).[7] Geschlecht wird dabei maßgeblich semantisch ausgedrückt, ohne grundlegend in die grammatikalischen Strukturen einer Sprache einzugreifen. In Sprachen ohne Geschlecht (z. B. Türkisch, Finnisch oder den meisten Gebärdensprachen) wird weder bei Nomina noch bei Pronomina ein Geschlecht ausgedrückt, bis auf wenige Ausnahmen etwa bei Verwandtschaftsbeziehungen (z. B. *sisko*, im Finnischen für „Schwester").

---

[1] Wir verwenden in diesem Artikel im weiteren Verlauf den Begriff *nicht-binär* als Oberbegriff für alle Geschlechtsidentitäten außerhalb des binären Schemas von Frau/Mann.

[2] In bspw. der Schweiz und Luxemburg gibt es derzeit noch keine Option für einen nicht-binären Geschlechtseintrag.

[3] Bei der Verwendung des Asterisks zwischen geschlechtlich binär eingeordneten Endungen handelt es sich um eine Version des so genannten *Gender-Gaps*. Das Schriftzeichen soll alle über Frauen und Männer hinausgehende Geschlechter symbolisieren und wird in der gesprochenen Sprache durch eine kurze Pause (genauer: einen Glottalplosiv, wie etwa auch an der Wortfuge in „Spiegelei") hörbar gemacht [25, S. 2 ff] [17, S. 27 ff]. Wir wenden in jedem Abschnitt dieses Artikels eine andere genderfaire Strategie an, um deren Vielfalt aufzuzeigen.

[4] Mit dem Oberbegriff „queer" verweisen wir auf jegliche Gruppenzugehörigkeit außerhalb des cis-heteronormativen Schemas, also jedwede Person, die sich innerhalb der LGBTQIA+-Community im weitesten Sinne verortet. Auch wenn nicht alle Personen aus dieser erweiterten Gruppe mit den gleichen Problemen im Bezug auf Sprachtechnologien zu kämpfen haben, gibt es doch eine erhöhte Sensibilität durch eine geteilte Erfahrung der Devianz von erwarteten Geschlechternormen (inklusive der damit verbundenen Sexualitätsnormen).

[5] Die Hauptautor*innen dieses Artikels konnten sich nicht auf eine konsistente Schreibweise für die Mehrzahl des eingedeutschten Wortes „Community" einigen, weshalb wir auf den nächstbesten Begriff ausweichen mussten, auch wenn dieser kein exaktes Synonym ist.

[6] Übersetzt aus dem Englischen „notional gender".

Historisch, und teils noch heutzutage, wird in deutschen Texten häufig maskulin gegenderte Sprache im Gegensatz zu *ent*genderter, genderfairer Sprache verwendet [32]. Diskussionen um die Sinnhaftigkeit sowie Wirkung von genderfairer Sprache füllen regelmäßig Kolumnen namhafter Publikationen, auch wenn unter Linguist*innen Konsens darüber herrscht, dass und wie sich Gender Bias in „natürlicher" bzw. von Menschen verwendeter Sprache auswirkt. Entsprechende Forschung begann in den 1970er Jahren [8] und diverse Studien, die die Verwendung des sogenannten „generischen Maskulinums" untersucht haben, legen nahe, dass diese Form nicht als „generisch" wahrgenommen wird, sondern einen klaren (männliche Assoziationen verstärkenden) Einfluss auf die hervorgerufenen mentalen Repräsentationen hat [4, 24, 36]. Dies hat sich im übrigen auch in üblichen Ausdrücken innerhalb der Mensch-Computer-Interaktion in der englischen Sprache bestätigt [12]. Entsprechend empfehlen die jeweiligen Forschens[8] sowie Autorens explizit geschlechterinklusive Sprache. Neben den Effekten auf mentale Repräsentation ist die Verwendung genderfairer Sprache auch im Interesse präziser Ausdrucksweise: wird nur das „generische" Maskulinum oder binär „inklusive" Sprache zur Beschreibung von Personen(gruppen) verwendet, ist unklar, wie diese Beschreibungen tatsächlich zu verstehen sind – es findet also ein Informationsverlust statt. So wird beispielsweise auf Spanisch oder Französisch von einer Gruppe von Menschen automatisch in der „generisch" maskulinen Form gesprochen, sobald sich auch nur ein Mann unter ihnen befindet, wobei durch die Verwendung von geschlechterinklusiver Sprache hier nuancierter formuliert werden kann.

Neben institutionellen Leitfäden [1, 17, 25] existieren diverse freiwillige Initiativen, die Informationen, Definitionen und Handreichungen rund um nicht-binär inklusive Sprache anbieten, etwa das *Nichtbinär-Wiki*[9] oder das Online-Wörterbuch *Geschickt Gendern*.[10] Derartige diverse Strategien werden jedoch bislang nicht umfassend in der Gesamtgesellschaft rezipiert, wie beispielsweise Spiel in Bezug auf digitale Systeme aufzeigt [37].

Eine weiter verbreitete und häufig diskutierte Strategie für genderfaire Sprache ist der sogenannte *Gender-Gap*, markiert durch einen Asterisk (wie in diesem Text, bspw. *Forscher\*innen*), Unterstrich (wie in den Danksagungen, bspw. *Forscher_innen*) oder Doppelpunkt (bspw. *Forscher:innen*) zwischen der maskulinen Form eines Nomens und dem femininen Suffix, um sich auf nicht-binäre Personen, Personen unbekannten Geschlechts oder gemischtgeschlechtliche Gruppen zu beziehen. Doch sind auch die Diskussionen rund um diese Strategie nach wie vor stark emotionalisiert und bleiben damit fortwährend Bestandteil gesellschaftlicher Auseinandersetzungen, wobei „[d]ie Argumente [...] sich in fast 50 Jahren nicht wirklich verändert [haben]" [19].

## 2.2 Von der „natürlichen" Sprache zur maschinellen – und zurück

Die häufige Verwendung des „generischen" Maskulinums in historischen Texten führt dazu, dass auf Basis der daraus konstruierten Korpora angelernte Sprach- und Übersetzungstechnologien diese exkludierende Form als Norm identifizieren und in Konsequenz als solche weiterführen und anwenden. Beispiele finden sich bei expliziten Übersetzungsanwendungen, die in der Ausgangssprache geschlechtsfrei formulierte Sätze stereotypen-belastet übersetzen (z. B. von *nurse* zu *Krankenschwester* anstelle von bspw. *Krankenpfleger\*in*). Nur durch dezidiertes Eingreifen, beispielsweise durch regelbasierte Mechanismen nach der automatisierten Übersetzung, explizites Gender-Tagging jeglicher geschlechtsbezogener Inflektionen (egal, welchen Bezug eine Sprache zu Geschlecht hat) oder eine explizite Nennung von Mehrfachübersetzungen kann einem derartigen Bias entgegengewirkt werden [35]. Doch nicht jede Anwendung von maschineller Übersetzung erfolgt transparent oder wird von Lesens wissentlich und bewusst verwendet: Häufig werden Online-Inhalte just-in-time und automatisch übersetzt, etwa Webseiten, Reviews oder Statusmeldungen. Auch die automatische Rechtschreibprüfung, die in vielen Textverarbeitungsprogrammen oder auch Browsern inkludiert ist, markiert geschlechtergerechte Formulierungen häufig als „falsch" und suggeriert damit, die Formulierung im „generischen" Maskulinum wäre korrekt(er) – auch weil im Duden derzeit nur eine limitierte Bandbreite an geschlechtergerechten Optionen verzeichnet ist.[11] Somit beeinflusst eine unausgewogene Sprachtechnologie, wie künftige Texte aussehen werden, und da viele Lern-Datensätze für Sprachverarbeitung Korpora aus Online-Quellen als Basis haben, beeinflusst die Sprachtechnologie wiederum auch, wie ihr künftiges Lernmaterial aussieht: Wir befinden uns also in einer immer weiter selbstverstärkenden Spirale.

---

[7] Wir folgen in diesem Text nicht strikt der im Englischen gebräuchlichen Unterscheidung zwischen „Sex" und „Gender"; es handelt sich bei dieser Dichotomie um sozial konstruierte, arbiträre Abgrenzungen, die ähnlich problematisch sind wie beispielsweise die zwischen „Körper" und „Geist" und viele andere (nur scheinbar natürliche) Gegensatzpaare [7].

[8] Bei der Endung -ens handelt es sich um eine geschlechtsneutrale Strategie, die eingeführt wurde, um auf eine Person im Deutschen ohne Geschlechtsreferenz verweisen zu können. Die Endung stammt dabei aus den Buchstaben, die sich in der Mitte des Wortes M*ens*ch befinden [23]. „Ens" kann dabei an einen Wortstamm angehängt werden oder auch eigenständig als Pronomen verwendet werden [23]. Singular bzw. Plural ergeben sich aus dem kontextuellen Bezug.

[9] https://nibi.space, abgerufen am 13. 4. 2022.
[10] https://geschicktgendern.de, abgerufen am 13. 4. 2022.
[11] Siehe https://www.duden.de/sprachwissen/sprachratgeber/Geschlechtergerechter-Sprachgebrauch, abgerufen am 13. 4. 2022.

Neben der rechtlichen Sicht [6] wurden in den letzten Jahren auch Sprachtechnologien beforscht und Stimmungsanalysen (Englisch: sentiment analysis) durchgeführt [11]. Lardelli etwa betrachtet Google Translate und DeepL, zwei häufig verwendete kommerzielle Systeme für maschinelle Übersetzung, und deren Herangehensweise bei Übersetzungen vom Englischen ins Italienische [29, 31]. Lardelli untersucht hierbei insbesondere das Zusammenspiel zwischen Konnotationen von Adjektiven und Berufsbezeichnungen. Dabei stellt er fest, dass es zu verschiedenen Fehlübersetzungen und Inkonsistenzen kommt, je nachdem, ob Adjektive oder Berufe eher männlich oder weiblich konnotiert sind [29, S. 48 ff]. Kombinationen von traditionell weiblich konnotierten Vornamen (*Martha* bzw. *Adele*) mit männerdominierten Berufsfeldern (z. B. *engineer*) führten zu schlechteren Übersetzungen als traditionell männlich konnotierte Vornamen (*Liam*) mit eher frauendominierten Berufen (z. B. *receptionist*). Dieses Verhalten konnte Lardelli in beiden Übersetzungsplattformen dokumentieren. Wachowiak et al. zeigen, dass Bias in Embeddings und Sprachmodellen nicht nur direkt in Nomina selbst vorhanden ist, sondern sich auch in Synonymen (gleichbedeutenden Begriffen), Antonymen (Gegenworten) und Hyperonymen (Oberbegriffen) niederschlägt [43].

Debiasing-Ansätze verschiedener Forschungsrichtungen versuchen auf unterschiedliche Arten, das beschriebene Problem in den Griff zu bekommen. Im Folgenden präsentieren wir exemplarisch einige Ansätze: Vanmassenhove et al. untersuchten Übersetzungen zwischen Englisch und mehreren anderen Sprachen und stellten fest, dass Tagging, also das Hinzufügen von kontextabhängigen Informationen, die Ergebnisse verbessert [42]; Moryossef et al. untersuchten das Sprachpaar Englisch–Hebräisch und stellten fest, dass Pre-Editing des Textes mit Zusatzinformation (etwa in Form von kurzen zusätzlichen Sätzen wie „she said") zu verbesserten Ergebnissen führt [30]; Stafanovičs et al. untersuchten ebenfalls Übersetzungen zwischen Englisch und mehreren anderen Sprachen; Tagging sowie Annotieren von Worten in den zielsprachigen Sequenzen (mittels F, M, N(euter), U(navailable)) lieferte auch hier verbesserte Ergebnisse [40]. Das „Entfernen" von Bias aus Word Embeddings resultierte ebenfalls in leichten Verbesserungen [18, 44]. Allerdings kommt dieser Ansatz mit neuen Ungenauigkeiten: Wer entscheidet, welche Position(en) in einer Relation zwischen Begriffen „neutral" oder „angebracht" ist? Diese Entscheidungen werden von subjektiven Vorannahmen und Einstellungen der Forschenden geprägt, die bislang oft nicht (ausreichend) reflektiert werden [10]. Auf dieser Meta-Ebene gibt es ebenfalls erste Forschungsarbeiten, insbesondere mit Blick auf die Beweggründe für Entscheidungen und Annahmen der Machine-Learning-Entwicklens [2].

In Summe lässt sich jedenfalls sagen, dass bisher verbreitete Ansätze oft technisch-mathematisch inspiriert sind, den weiteren wissenschaftlichen Kontext sowie die gesellschaftlichen Wechselwirkungen ignorieren oder nicht ausreichend berücksichtigen, die Betroffenen kaum bis gar nicht einbeziehen und Forschens ihre Annahmen und Ziele selten explizit bzw. klar formulieren [10]. Weiters setzt die existierende Forschung weiterhin fast ausschließlich ein traditionelles Verständnis von Geschlecht als binär, unveränderlich, unveränderbar und biologisch [26] voraus [41].

Dementsprechend identifizieren wir folgende Forschungslücke: Alle betroffenen Gruppen – nämlich nicht-binäre und queere Personen, Übersetzens und Expertens der maschinellen Übersetzung – an einen Tisch bringen; darüber sprechen, was die Probleme für die verschiedenen Betroffenen sind; und den Entwicklens und Übersetzens, denen das Bewusstsein noch fehlt, vermitteln, warum „langsame" Lösungen (wie z. B. neue Datensätze als Basis zu finden oder zu erzeugen) nicht ausreichen. Ansätze, bei denen die von Erasure (engl. für Löschung oder Tilgung; bewusste oder unbewusste Unsichtbarmachung marginalisierter Identitäten durch Auslassung), Re-Stigmatisierung und Bias Betroffenen kontinuierlich ihre eigene Benachteiligung erfahren, weiter erdulden und die Thematik (wiederholt) erklären [9] müssen („trickle-down justice"), sollten dahingehend beim Finden von potenziellen Lösungen streng vermieden werden.

## 3 WORKSHOP

Der Workshop fand an drei aufeinander folgenden Tagen im Herbst 2021 an der TU Wien statt. Basierend auf Prinzipien des Participatory Action Research [27] wurden unterschiedliche relevante StakeholderNinnen[12] eingeladen, um an Gestaltungsvoraussetzungen für genderfaire maschinelle Übersetzung zu arbeiten. VertreterNinnen der nicht-binären und queeren Gemeinschaften, ÜbersetzerNinnen und ExpertNinnen für maschinelles Übersetzen begegneten einander in abwechselnden Kleingruppen- und Plenareinheiten. Dabei wurden die Kleingruppen themen-, expertise- und aufgabenabhängig durchmischt. Um die Veranstaltung möglichst zugänglich zu gestalten und unterschiedliche marginalisierte Personengruppen zu inkludieren [14], wurde bei der Organisation besonderes Augenmerk auf Zugänglichkeit gelegt: Der Ort der Veranstaltung war durch öffentliche Verkehrsmittel gut erreichbar, es wurden Nahrungsmittelunverträglichkeiten für die (regionale, nachhaltige) Verpflegung berücksichtigt, und es wurde ein tagesaktuelles COVID-19-Hygienekonzept entsprechend den Vorgaben der TU Wien und der pandemischen Gesamtsituation kommuniziert und befolgt (tägliche Kontrolle des Gesundheitsstatus (geimpft, genesen, getestet) beim Betreten des Gebäudes; Tragen von FFP2-Masken in den Gängen; Contact-Tracing für die Veranstaltung im Besonderen sowie für das Gebäude insgesamt).

### 3.1 TeilnehmerNinnen

Da es sich bei genderfairer Sprache um ein emotional aufgeladenes Thema handelt [19], war es hilfreich, dass in der Gruppe der OrganisatorNinnen bereits nicht-binäre und queere Gemeinschaften, MT-ExpertNinnen und ÜbersetzerNinnen vertreten waren. Potenzielle TeilnehmerNinnen konnten über unterschiedliche Online-Gruppen und E-Mail-Newsletter aktivistischer Vereine, persönliche Kontakte und allgemeine Ausschreibungen in professionellen Kontexten (Aushänge, direkte Ansprache, …) gewonnen werden. Insgesamt nahmen am Workshop 21 Personen zusätzlich zu den OrganisatorNinnen teil, von denen einige mehr als einer der genannten Personengruppen zugehörig waren (beispielsweise queere ÜbersetzerNinnen oder ÜbersetzerNinnen mit MT-Expertise), auch wenn die Gruppe der Community-VertreterNinnen sehr viel kleiner war.

Die TeilnehmerNinnen kamen maßgeblich aus Österreich, bis auf einige ExpertNinnen, die aus Deutschland bzw. der Schweiz

---

[12] In diesem Abschnitt verwenden wir die Sylvain-Konvention, welche ein komplettes Sprachsystem mit Deklination der jeweilig zugehörigen Artikel und Pronomina vorschreibt. Es handelt sich dabei um ein geschlechterinklusives System, mit dem eine zusätzliche, explizit neutrale Form eingeführt wird [16].

angereist kamen. Alle TeilnehmerNinnen (und OrganisatorNinnen) waren *weiß*,[13] was den Gültigkeitsanspruch unserer Ergebnisse dahingehend auf ein Geschlechterverständnis mitteleuropäischer Mehrheitsgesellschaften reduziert. Etwa vier TeilnehmerNinnen und OrganisatorNinnen identifizieren sich zudem auch als der Behindertencommunity zugehörig.

Um den TeilnehmerNinnen zu ermöglichen, sich im Vorfeld ausreichend mit dem Thema des Workshops auseinanderzusetzen und ggf. Nachfragen zu stellen (und somit mit ausgeglichenerem Vorwissen in den Workshop starten zu können), wurde ein Handout zur Vorbereitung erstellt und vorab ausgeschickt, in dem einige genderfaire Strategien für deutschsprachige Texte erläutert sowie weitergehende Leseempfehlungen angegeben wurden.[14]

## 3.2 Methoden und Ergebnisse

Wie bereits erwähnt, wurde zwischen Kleingruppen- und Plenaraktivitäten regelmäßig abgewechselt. Auch die Zusammensetzung der Kleingruppen wurde durchmischt: Für manche Aktivitäten blieben die jeweiligen StakeholderNinnen-Gruppen unter sich, andere wurden in gemischten Gruppen absolviert, wobei auch die OrganisatorNinnen an den jeweiligen Kleingruppen in ihren unterschiedlichen Funktionen und Zugehörigkeiten teilnahmen. Für jede Aktivität erhielten die TeilnehmerNinnen zusätzlich zu einer verbalen Einleitung auch eine schriftliche Erklärung sowie Hinweise zu verfügbaren oder vorgeschlagenen Materialien zur Bearbeitung. Am Ende jeden Tages wurde im Plenum eine gemeinsame Zusammenfassung erstellt. Der erste Tag diente dem gegenseitigen Kennenlernen sowie der Einführung der unterschiedlichen Positionen zur Problematik genderfairer maschineller Übersetzung sowie der jeweiligen Imaginierung eines Idealzustandes. Der zweite Tag war der Entwicklung und Erprobung unterschiedlicher Strategien gewidmet, um dann am dritten Tag weiteres Vorgehen abzusprechen und mit der öffentlichen Podiumsdiskussion abzuschließen. Im Folgenden gehen wir auf jede der unterschiedlichen Einheiten, die verwendeten Methoden sowie die Ergebnisse genauer ein.

*3.2.1 Ankommen und Kennenlernen.* Bei ihrer Ankunft wurden die TeilnehmerNinnen auf die Möglichkeit hingewiesen, ihre Zugehörigkeit zu den Personengruppen sowie ihre Pronomina mittels verschiedenfarbiger Aufkleber auf ihren Namensschildern anzuzeigen. Nach der allgemeinen Begrüßung durch das Organisationsteam folgte ein Kennenlernen durch soziometrisches Aufstellen – die TeilnehmerNinnen wurden gebeten, sich als Antwort auf verschiedene Fragen im Raum zu positionieren. Die Fragen waren dabei teils entsprechend dem Workshop-Thema (beispielsweise „Kenntnisse zu genderfairer Sprache" oder „Erfahrung mit MT"), teils unabhängig davon (etwa, wie zum Workshop angereist worden war). Durch diese Interventionen konnten die TeilnehmerNinnen binnen recht kurzer Zeit Bezüge und Beziehungen zueinander herstellen und schlugen auch selbst eigene Fragen vor.

---

[13] Wir setzen *weiß* in diesem Text in Anlehnung an [3, S. 13] kursiv. Hintergrund ist, auf die soziale Konstruktion des *Weiß*-Seins hinzuweisen und die Kategorie gleichzeitig abzugrenzen vom Widerstands- und Ermächtigungspotential der (Selbst-)Bezeichnung und Kategorie „Schwarz" bzw. „of Colour".
[14] Eine modifizierte englische Version, die für eine Aufgabe im Rahmen eines Hackathons erstellt wurde, ist hier verfügbar: https://www.goethe.de/resources/files/pdf237/introductory-handout-for-challenge-2.pdf, abgerufen am 13. 4. 2022.

*3.2.2 Problemstorming.* Zu Beginn war es notwendig, möglichst sicherzustellen, dass alle TeilnehmerNinnen eine zumindest grundlegend geteilte Auffassung in Bezug auf das Verständnis der Problemlage entwickelten. Entsprechend startete der Workshop mit einer Anaylse der aktuellen Ist-Situation hinsichtlich genderfairer Sprache und maschineller Übersetzung. In dieser ersten Einheit wurden gemischte Kleingruppen mit drei bis vier TeilnehmerNinnen gebildet, so dass möglichst jede StakeholderNinnen-Gruppe in jeder Kleingruppe vertreten war. Ziel war in diesem Schritt ein Austausch über Erfahrungen, Bedürfnisse und potenzielle Schwierigkeiten der genderfairen (maschineller) Übersetzung. Zur Anregung erhielten die TeilnehmerNinnen ein Handout, das Beschreibungen von nicht-binären Personen und maschinelle Übersetzungen der Beschreibungstexte enthielt. Die Verwendung von singular they in manchen Originaltexten führte zu genau dem Problem, das im Rahmen des Workshops bearbeitet werden sollte: Fehlerhafte Übersetzungen zu pluralem „sie" und Misgendering der beschriebenen Personen. Auf diesen Umstand wurde bereits vor der Ausgabe der Handouts hingewiesen, sodass vor allem queere und nicht-binäre TeilnehmerNinnen die Option hatten, diesen Teil der Aktivität auszulassen (da eine falsche geschlechtliche Zuschreibung oft an eigens erlebte, teils schmerzhafte, Verletzungen und Diskriminierungserfahrungen erinnert).

Im Rahmen dieser Kleingruppenarbeit stellte sich für viele heraus, dass Vielförmigkeit und Potenzial für Kreativität beim Entgendern im Deutschen zu Schwierigkeiten bei der Durchsetzung der *einen*, „richtigen", „besten" Strategie führen. Demnach kamen die Anwesenden zu der Übereinstimmung, dass individuelle Personen im Idealfall nach ihren Pronomina und bevorzugter genderfairer Bezeichnung gefragt werden sollten. Dies kann sich aber in vielen Kontexten als schwierig bis unmöglich gestalten, insbesondere in maschinellen Vorgängen oder gemischten Gruppen mit potenziell unterschiedlichen Präferenzen sowie in Fällen, in denen eine Person die Bandbreite an möglichen Übersetzungen der eigenen Pronomina in eine bestimmte Zielsprache nicht kennt. Durch diesen hohen Grad an Kontextabhängigkeit wurde schnell klar, dass es nicht „die eine" Strategie für alle Situationen geben kann. Dementsprechend entwickelte sich die Redewendung „One size fails all" (angelehnt an und im Kontrast zu „One size fits all") im Laufe des Workshops regelrecht zu einem geflügelten Wort. Weitere Aspekte, die laut den TeilnehmerNinnen des Workshops zur Problemkomplexität beitragen, liegen darin, dass die Auswahl einer genderfairen Sprachstrategie auch unterschiedliche sozio-technische Aspekte mitberücksichtigen muss. Beispielsweise sollten derartige Strategien auch bezüglich inklusiver Lesbarkeit und Verständlichkeit auch für sprachlernende und behinderte Personen adäquat funktionieren. Von eher wirtschaftlich-industrieller Seite wurde zudem Inkonsistenz als Teilproblem erkannt, weil hier übergeordnete technische Systeme, bspw. bei der Optimierung von Suchmaschinenrankings, noch wenig verbreitete Sprachformulierungen eher nachrangig behandeln. Zudem wurde explizit von ExpertNinnen der maschinellen Übersetzung aufgeworfen, dass fehlende Beispieltexte und Trainings-Korpora derzeit das größte Hindernis für das Erlernen von automatisierten genderfairen Übersetzungsstrategien darstellen.

Bereits in dieser Phase wurden erste Ideen für Lösungsmöglichkeiten entwickelt. Darunter befand sich beispielsweise der Vorschlag, NutzerNinnen die Möglichkeit zu geben, einen Text mit einer gewählten Strategie genderfair zu übersetzen, oder (als einen ersten Schritt) Übersetzungen von nicht genderfairem Deutsch in genderfaires Deutsch („intralinguale Übersetzung") zu erkunden. Außerdem wurde festgestellt, dass es nicht ausreichen wird, ausschließlich auf grammatikalisch-syntaktischer Ebene zu arbeiten, um Inklusivität zu erreichen, sondern dass auch althergebrachte Rollenbilder und stereotype Ansichten die geschlechtsbezogene Inklusivität eines Textes stark beeinflussen. Dahingehend gestaltet sich die Problemlage zusammenfassend als hochkomplex, mit einem hohen Grad an Kontextbezogenheit und einer fehlenden ausreichenden Datengrundlage für moderne Ansätze der maschinellen Übersetzung.

*3.2.3 Utopienstorming.* Nach der Ausarbeitung der Problemlage wurden die TeilnehmerNinnen dazu aufgefordert, je nach ihrer Expertise Zielvorgaben in der Form von Utopien auszuarbeiten. Dies hatte einerseits zum Zweck, einen erwünschten Idealzustand zu definieren, um später Lösungsansätze vom Ist-Zustand hin zum imaginierten Ideal ausarbeiten zu können. Andererseits war es uns aber auch wichtig, den ersten Tag mit einem positiven, optimistischen Blickwinkel enden zu lassen, der nicht nur von der Unzulänglichkeit der jetzigen Situation gespeist war.

Diese Aktivität fand demnach community-intern statt, also in den drei Gruppen der nicht-binären und queeren Personen, ÜbersetzerNinnen und technischen ExpertNinnen. Personen, die mehreren dieser Gruppen zugehörig waren, wählten dabei nach eigenem Ermessen, welche Utopien sie mitentwickeln wollten. Die Vorgabe war dabei, Hoffnungen, Wünsche, Träume bezüglich (maschineller) genderfairer Übersetzung zu sammeln. Den Gruppen wurde freigestellt, wie sie ihre Ergebnisse bzw. Diskussionen dokumentieren. Sie wurden dabei auch explizit zu kreativen Herangehensweisen ermutigt, entsprechendes Material zum „Basteln" wurde zur Verfügung gestellt. Nach anfänglichen Vorbehalten gegenüber der Anregung zum Basteln entstanden auch abstraktere Visualisierungen der jeweiligen Gedankengänge in den Gruppen, welche den utopischen sowie inklusiven und vielfältigen (weil vielfarbigen) Charakter der Überlegungen darstellen sollte: Leonda, the Gender Avenger wurde primär als Maskottchen beschrieben, während das eierlegende Wollmilch-Wir-sind-Wir humorvoll auf den Wunsch nach „einer Lösung für alles" verweist und dabei auch eine Visualisierung eines möglichen weiteren Prozesses (Stufenmodell) beinhaltet (siehe Abbildung 1).

ÜbersetzerNinnen und MT-ExpertNinnen wünschten sich vor allem Standards und möglichst klare Vorgaben (mit einem gewissen Grad an Flexibilität). Die Standardisierung würde einerseits helfen, bei LeserNinnen und KundNinnen Akzeptanz aufzubauen, und andererseits wären die daraus resultierenden wiederkehrenden Muster in Texten hilfreich für maschinelle Übersetzungstechnologien. Die eingeforderte Flexibilität sollte ermöglichen, auch jene Personen(gruppen) sichtbar zu machen, die von der Reduktion und Vereinfachung, die Standards mit sich bringen, nicht berücksichtigt würden. Außerdem wünschte sich diese Gruppe eine zentrale Anlaufstelle mit gewisser Autorität, etwa eine Hotline, die bei Fragen zu genderfairer Sprache kontaktiert werden könnte.

Die Utopien, Wünsche und Hoffnungen der queeren und nichtbinären TeilnehmerNinnen verweisen vor allem auf eine harsche Realität bzw. aktuelle Situation, da sie sich maßgeblich darum drehten, dass Respekt, „einfach sein können" und Rücksichtnahme eingefordert wurden, was alle Beteiligten (auch der anderen Gruppen) als „eigentlich eine Selbstverständlichkeit" bewerteten (siehe dazu auch [38]). Allerdings wurde hier auch deutlich, dass eine einheitliche Lösung von der Community letztlich nicht gewünscht ist, da sie der Flexibilität, oft dynamischen Entwicklung und Vielfalt in der individuellen Selbstbestimmung zuwiderläuft.

In der abschließenden Plenareinheit wurde schließlich besprochen, welche politischen, rechtlichen und gesellschaftlichen Rahmenbedingungen notwendig wären, um die Nachfrage nach – und damit Akzeptanz von – genderfairer Sprache zu fördern. Als eine Möglichkeit wurden Anreize für Unternehmen angesprochen, damit diese zunehmend genderfaire Sprache verwenden. Zusammenfassend wurde in der Erstellung der Utopien also deutlich, dass Bedürfnisse nach Flexibilität und individueller Ausdrucksmöglichkeit mit denen nach Standardisierung und Eindeutigkeit verhandelt werden müssen.

*3.2.4 Anwendungsbeispiel: Fiktive Charaktere genderfair beschreiben.* Als Start in den zweiten Tag des Workshops wurden nach individueller Expertise gemischte Kleingruppen gebildet und deutsche Beschreibungen von fiktiven Charakteren, wie Arielle, die Meerjungfrau, Peter Pan oder Pippi Langstrumpf, ausgeteilt. Die Aufgabe für die TeilnehmerNinnen bestand darin, die Texte genderfair umzugestalten. Es war den jeweiligen Gruppen freigestellt, nur eine Strategie zu wählen, oder verschiedene auszuprobieren; auch die Wahl der konkreten genderfairen Sprachform war völlig freigestellt. Zur Hilfestellung konnte das vorher verteilte Handout verwendet werden.

Einige Gruppen wählten die Strategie, Pronomina weitestgehend zu vermeiden und stattdessen auf direkte Namensnennung oder Passiv-Konstruktionen auszuweichen und Nomina durch deren Pluralformen oder neutrale Formen zu ersetzen, wie z. B. *Meerwesen* statt *Meerjungfrau*. Zwei Gruppen verwendeten das genderfaire Dey-e-System.[15] Manche TeilnehmerNinnen, die mit der Nutzung genderfairer Sprache bereits vertraut waren, fanden es einfacher, die Texte komplett neu zu schreiben statt Satz für Satz oder Wort für Wort zu übersetzen. Bei einer reinen „Übersetzung", stellten sie fest, bestehe die Gefahr, geschlechtsspezifische Bezeichnungen zu übersehen. Seltene oder weniger offensichtlich geschlechtsspezifische Begriffe (wie eben beispielsweise *Meerjungfrau*) zu erkennen und passend zu ersetzen ist nicht nur ein Problem für Menschen, sondern auch ein offenes Forschungsthema im Bereich der MT. Durch die eigenständig durchgeführten Experimente konnten die WorkshopteilnehmerNinnen sich auch noch einmal mit unterschiedlichen Systemen vertraut machen, diese besser im Praxisgebrauch kennen lernen und so individuelle Sicherheit im Umgang mit genderfairer Sprache in für sie teilweise neuen Systemen gewinnen.

*3.2.5 Strategienstorming.* Nachdem in den vorherigen Einheiten der aktuelle Stand im Bezug auf genderfaire Sprache und Sprachtechnologien sowie mögliche Ziele und Utopien geklärt werden konnten, wurde als Nächstes darüber gesprochen, welche Wege von der derzeitigen Problemlage hin zu den besprochenen Utopien möglich wären. Wichtig war in dieser Einheit, dass die vier gemischten Kleingruppen kein großes Augenmerk auf die Machbarkeit ihrer Vorschläge für genderfaire Übersetzung und MT legten (daher hier die Bezeichnung des Abschnittes als „-storming").

---

[15] Wie z. B. *einey gute Arzte*; siehe dazu https://geschlechtsneutral.net/dey-e-system-archivseite/, abgerufen am 13. 4. 2022.

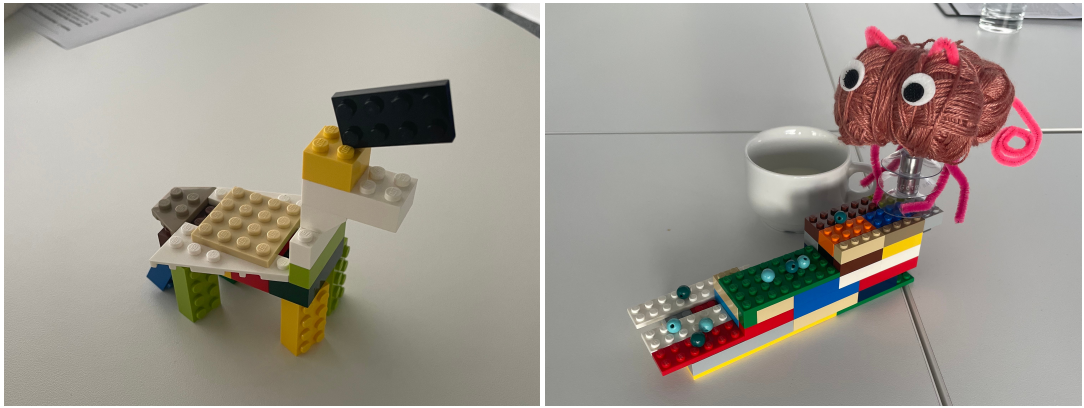

**Abbildung 1: Leonda, the Gender Avenger, und das eierlegende Wollmilch-Wir-sind-Wir.**

Ein Vorschlag der TeilnehmerNinnen war, einen Standard für genderfaire Sprache zu finden, um Sicherheit bezüglich grammatikalischer Korrektheit zu schaffen und die Aussprache einfach und einheitlich gestalten zu können. Dieser Weg könnte aus dieser Perspektive auch dazu führen, dass Akzeptanz und Verwendung in der breiteren Gesellschaft und in Institutionen gefunden wird, wie z. B. im Duden, bei MedienvertreterNinnen und Behörden. Während solch eine Standardisierung die Akzeptanz geschlechtsneutraler Sprache erhöhen könnte, widerspräche sie gleichzeitig dem Verständnis einer sich laufend verändernden und entwickelnden Sprache, dem Verständnis von Geschlecht als etwas grundsätzlich Fluides und Veränderliches sowie unserer Vorstellung von genderfairer Sprache, für die ebenfalls laufend neue Strategien gefunden und entwickelt werden. Als Alternative wurden flexible Leitlinien vorgeschlagen, die einen gewissen Grad an Standardisierung ermöglichen, aber gleichzeitig Möglichkeiten bewahren, um die Sprache zu personalisieren und weiterzuentwickeln.

Eine Reduktion der – vor allem im Deutschen – großen Vielfalt an genderfairen Sprachstrategien würde auch die weitere Entwicklung von genderfairer automatisierter Übersetzung immens vereinfachen. Genderfaire Beispieltexte und Datensätze wären schneller in größerer Zahl verfügbar. In diesem Zusammenhang wurde auch vorgeschlagen, Beispieltexte regelbasiert zu erstellen (siehe auch die Idee von intralingualer Übersetzung im Rahmen des Problemstorming), und ein neues Berufsfeld des genderfairen Pre- und Post-Editing angedacht. Personen mit dieser Aufgabe würden Texte für MT adäquat aufbereiten (bspw. durch Tagging) und/oder im Nachgang die Übersetzungen kontrollieren und nachbearbeiten. Im ersteren Fall können MT-Modelle durch das Tagging eigenständig genderfaire Strategien übernehmen, im zweiteren wird die Qualität durch menschliche Intervention sichergestellt. Die beiden Ansätze lassen sich natürlich auch kombinieren.

*3.2.6 Strategising.* In den gleichen Gruppen wie in der vorhergehenden Einheit wurden die bisher entwickelten und überlegten Strategien auf ihre Sinnhaftigkeit, Übersetzbarkeit und sozio-technische Machbarkeit hin überprüft.

Eine Idee, die aus diesen Diskussionen entstand, war die Entwicklung eines *Stufenmodells*, anhand dessen der eigene Sprachgebrauch eingeschätzt und Fortschritt in Richtung genderfairer Sprache erkannt werden könnte. Das Stufenmodell beginnt mit der linguistischen Inklusion von Frauen, etwa in Form des Binnen-I (*LeserInnen*) oder des „generischen" Femininums (*Leserinnen*). Die zweite Stufe inkludiert nicht-binäre Personen, indem der Gender-Gap verwendet wird (siehe auch Abschnitt 2.1), wobei hier von den MT-ExpertNinnen auf die Mehrfachbelegung des Asterisks in technischen Sprachen und damit einhergehenden Unklarheiten und potenziellen Probleme verwiesen wurde. Die dritte und letzte Stufe des Modells zielt darauf ab, ein Outing des Geschlechts einer Person zu vermeiden, was durch genderneutrale Ansätze möglich ist. Manche TeilnehmerNinnen äußerten hierbei eine Präferenz für die *-ens*-Strategie (aus M**ens**ch; z. B. *Lesens*). Diese Strategie würde vor allem aus technischer Hinsicht einige Schwierigkeiten vermeiden, da sie keine Sonderzeichen verwendet und Wörter nicht durch Suffixe verlängert (relevant etwa bei Zeichenbeschränkungen in Interface-Designs). Außerdem bleibt die Lesbarkeit bestehen und die Aussprache muss nicht zusätzlich geklärt werden. Dagegen steht allerdings, dass diese Form sehr nahe an den deutschen Genitiv kommt und dadurch Verwechslungsgefahr besteht; außerdem ist in der Strategie bisher keine explizite Pluralform vorgesehen, was zu einem gewissen Maß an Informationsverlust führt. Zudem würde dadurch eine dezidiert geschlechts*neutrale* Strategie als geschlechts*inklusiver* Ansatz verwendet, was der ursprünglichen Konzeptionalisierung widerspricht. Eine weitere Gruppe argumentierte für die Verwendung der Sylvain-Konventionen [16], die ein viertes Geschlecht, das *Liminalis*, einführen. Es handelt sich hierbei somit um eine geschlechterinklusive Strategie, die Kontext erhält, direkte Übersetzung von Äquivalenten erleichtert und eine konsistente genderfaire Sprache ohne Informationsverlust ermöglicht. Allerdings ist auch hier Vorsicht angeraten: Nicht jedes singular they wird korrekt als *nin* übersetzt.

In dieser Einheit wurden auch Kriterien für die Auswahl einer Strategie besprochen: Verwendbarkeit, Zugänglichkeit, Allgemeingültigkeit, Lesbarkeit, Verständlichkeit und Akzeptanz. Das Stufenmodell kann als mehrstufiger Anpassungsprozess in MT implementiert werden, um einerseits genderfaire Übersetzung zu ermöglichen, aber andererseits auch um unterschiedliche Wünsche mit AuftraggeberNinnen strukturiert besprechen zu können. Die genannten Kriterien können dann bei der Auswahl der spezifischen genderfairen MT-Strategie als Entscheidungshilfe herangezogen werden.

Die wiederholt auftauchende Ambivalenz zwischen inklusiver und geschlechtsneutraler Sprache zeigte sich auch in einer Umfrage, die am letzten Tag des Workshops in einer Plenareinheit durchgeführt wurde. Gefragt nach ihrer allgemeinen Präferenz, teilten sich die Antworten der TeilnehmerNinnen zu gleichen Teilen zwischen geschlechtsinklusiven und geschlechtsneutralen Strategien auf. Mit Blick auf dieses Ergebnis wurde vorgeschlagen, ein hybrides Modell zu entwickeln, das geschlechtsinklusive und -neutrale Sprache zu einem *genderfairen* Ansatz verbindet.

*3.2.7 Gegenseitige Bedarfsabfrage.* Die letzte Einheit des Workshops war darauf ausgelegt, spezifische Anregungen für weitere Entwicklungen zu sammeln. Die im Workshop gewonnene Dynamik sollte genutzt werden, um klare Vorschläge zu generieren, die die TeilnehmerNinnen und organisierenden ForscherNinnen mitnehmen konnten. Dabei wurden die Bedürfnisse der Personengruppen und ihre Wünsche an die jeweils anderen ExpertNinnen abgefragt und, im Sinne partizipativer Forschung [27], klar darauf verwiesen, dass diese Bedürfnisse nicht nur in Forschungskontexten, sondern auch in der Praxis bzw. im Rahmen aktivistischer Bestrebungen geäußert werden sollten.

TeilnehmerNinnen aus den **nicht-binären und queeren Gemeinschaften** äußerten den expliziten Wunsch, dass sich ÜbersetzerNinnen und Machine-Translation-ExpertNinnen aktiv als Allys, also als nicht-selbst-betroffene UnterstützerNinnen in der Sache, positionieren. Insbesondere ÜbersetzerNinnen wurden dabei als mögliche BrückenbauerNinnen zwischen der Mehrheitsgesellschaft und nicht-binären bzw. queeren Gruppen gesehen. Insgesamt wurde auch die trans- und interdisziplinäre Zusammenarbeit und der generell respektvolle Umgang innerhalb des Workshops lobend hervorgehoben. Während des Workshops kam es mehrfach zu Situationen, in denen ÜbersetzerNinnen und MT-ExpertNinnen Ängste dazu äußerten, dass ihre Versuche, inklusiv(er) zu handeln und zu sprechen, als „falsch" aufgenommen werden könnten.[16] Die VertreterNinnen der Community federten diese Ängste ab, indem sie mehrfach darauf hinwiesen, dass es darum gehe, überhaupt einmal irgendwo anzufangen, und nicht darum, gleich von Beginn an alles richtig zu machen. Und tatsächlich hat diese emotionale Arbeit[17] Früchte getragen und zu folgender Erkenntnis geführt: „Es geht hier um die Bedürfnisse von Menschen, nicht nur darum, die geilste Lösung zu implementieren." Für weitere künftige Arbeiten wurde der Wunsch formuliert, dass die Einbindung von (Selbst-)VertreterNinnen mit Rücksicht auf ihre Bedürfnisse, Erfahrungen und Wünsche fortgeführt werden sollte, ohne dabei anzunehmen, dass einzelne Personen für alle Individualperspektiven gesamtverantwortlich sein können.

Aus Sicht der **ÜbersetzerNinnen** ist eine eindeutige Zuweisung von Entsprechungen aus anderen Sprachen essenziell, unabhängig von der konkreten Strategie zu geschlechterinklusiver oder geschlechtsneutraler Formulierung. Informationen des Ausgangstextes sollten im Zieltext nicht verloren gehen, was in manchen Texten gegen eine geschlechtsneutrale Strategie spricht, und die Formen müssen eine kontextgetreue Formulierung ermöglichen. Hierbei gilt es vorrangig, verschiedene Strategien konkret über verschiedene Sprachtypen und Sprachpaare hinweg auf Übersetzbarkeit zu testen. In jedem Fall wäre für die Sprachindustrie ein Standard bzw. sogar ein „genderfaires Gütesiegel" von Vorteil, da dies Akzeptanz und Einhaltung neuer Sprachnormen erleichtern würden. Als Grundlage für weitere Arbeit wünschten sich die VertreterNinnen aus dem Übersetzungsbereich eine klare Aufstellung zur Verwendung genderfairer Sprache, sozusagen einen Schummelzettel („Cheat Sheet"), der im Übersetzungsprozess konsultiert werden kann.

Die **ExpertNinnen der maschinellen Übersetzung** formulierten als Hauptkriterium, dass die konkreten Einschränkungen der Umsetzbarkeit berücksichtigt werden sollten, z. B. limitierte Zeichenanzahl bzw. Feldlängen in bestimmten Kontexten wie BenutzerNinnenoberflächen, bereits existierende Belegung von bestimmten Zeichen mit konkreten Funktionen, besonders beim Asterisk, und explizite Klarheit über die zu verwendende Sprache in allen Kontexten. Für die maschinelle Übersetzung stellt der Mangel an verfügbaren Textbeispielen und Korpora ein zentrales Problem dar. Ein möglicher Lösungsansatz wäre auf Basis hybrider Methoden – einschließlich inzwischen weniger gebräuchlicher regelbasierter Verfahren – Textbeispiele zu generieren. Als Alternativlösung wurde eine intralinguale Übersetzung aus dem Deutschen in genderfaires Deutsch vorgeschlagen. Eine flexiblere Gestaltung maschineller Übersetzung wurde angedacht, in welcher NutzerNinnen die Wahl der genderfairen Strategie für den Zieltext selbst bestimmen können bzw. die Systeme kontextbasiert sinnvolle Vorschläge anbieten.

Insgesamt wurde im Rahmen der drei Workshoptage also die Problematik genderfairer automatischer Übersetzung aus unterschiedlichen Perspektiven verhandelt, die jeweiligen Zielvorstellungen entworfen und dargestellt und darauf aufbauend gemeinsam Lösungsstrategien erarbeitet.

## 4 (PODIUMS-)DISKUSSION

Die Ergebnisse des Workshops wurden schließlich am Ende des letzten Workshoptages noch in einer Podiumsdiskussion besprochen und einer interessierten Öffentlichkeit präsentiert. In der Zusammenfassung der dort diskutierten Aspekte sehen wir einen Ansatz zur partizipativen Bedeutungsbildung aus vorher kollaborativ erarbeiteten Ergebnissen [39]. Die Podiumsdiskussion bot zusätzlich den Anwesenden die Möglichkeit, Fragen zu stellen und selbst noch weiteren Input zu liefern. Die Veranstaltung, die am Erste Campus in Wien stattfand, wurde von einem Freiwilligen aufgezeichnet und geschnitten und im Anschluss auf YouTube[18] veröffentlicht. Die automatisch erstellten Untertitel wurden als qualitativ unzureichend eingeordnet und im Nachgang durch Teile des Forscherne-Teams[19] und weitere Freiwillige ausgebessert. Wie auch beim Workshop fanden auch bei der Podiumsdiskussion Kontrollen zum Gesundheitsstatus am Eingang statt. Anwesenderne wurden gebeten, Kontaktverfolgungsformulare auszufüllen.

---

[16] Zitat aus der Diskussion dazu: „… und dann handeln wir uns einen Shitstorm ein."
[17] Zum Begriff im Kontext von Design siehe [5].

[18] https://www.youtube.com/watch?v=RvCG5cL5ZSI, abgerufen am 13. 4. 2022.

Nach einer kurzen Vorstellung des Projektrahmens und einer Präsentation der Workshop-Ergebnisse diskutierten Tinou Ponzer (damals Vize-Obmensch vom Verein Intergeschlechtlicher Menschen Österreich (VIMÖ)), Rhonda D'Vine (Verein nicht-binär (VENIB)), Tristan Miller (Österreichisches Forschungsinstitut für Artificial Intelligence (OFAI)), Bettina Schreibmaier-Clasen (UNIVERSITAS Austria, Übersetzerin) und Anita Wilson (Kaleidoscope) die daraus folgenden Konsequenzen aus ihrer jeweiligen Perspektive.

### 4.1 Umgang mit genderfairer Sprache in Beruf, Ehrenamt und Alltag

Ein wichtiger Aspekt zu Beginn der Diskussion war, dass Unternehmen zwar an inklusiven Strategien interessiert sind, sie aber oft noch das „generische" Maskulinum verwenden, also durch grammatikalisch ausschließlich männliche Formen auf Personengruppen und Individuen unbekannten Geschlechts verweisen wollen. Dies wird teilweise mit der Annahme begründet, dass Strategien für genderfaire Sprache die Lesbarkeit von Texten beeinträchtigen könnten und sich damit negativ auf die Klickzahlen von Webseiten auswirken würden. Zum Beispiel benachteiligt Search Engine Optimisation (SEO) Webseiten, die mit genderfairen Strategien arbeiten, da Benutzerne bei der Suche im Internet mehrheitlich das „generische" Maskulinum verwenden (sicherlich auch, weil eben umgekehrt die meisten Webseiten dies tun; ein Teufelskreis).

Übersetzerne sind sich bewusst, dass sie genderfaire und im Hinblick auf weitere gesellschaftliche Unterdrückungsmechanismen wie Rassismus, Klassismus, Ableismus (Diskriminierung von behinderten, neurodivergenten sowie chronisch kranken Personen) usw. inklusive Sprache verwenden sollten. Oftmals ist beim Thema Inklusion von Seiten der Auftraggeberne jedoch aus Mangel an Hintergrundwissen und Awareness nur die Rede von der sprachlichen Gleichbehandlung von Frauen, weswegen es (aus der Perspektive unserer Arbeit insbesondere) noch Aufklärungs- und Diskussionsbedarf zum Thema „nicht-binär" gibt. Weiters kann sich die Übersetzung schwierig gestalten, wenn unbekannt ist, was de Autore beabsichtigt hat (ob z. B. das spezifizierte Geschlecht relevant oder eine freie Übersetzung erlaubt ist).

Tinou Ponzer bemerkte im Rahmen der Podiumsdiskussion zudem, wie bei der Mitarbeit am Buch „Inter*Pride" [21], welches sich ausführlich mit individuellen Erfahrungen zu Intergeschlechtlichkeit beschäftigt, die Übersetzungen von nicht deutschen Interviews und Beiträgen von inter* Aktivisterne ein starkes Proofreading seitens der Editorne benötigten, da die professionellen Übersetzerne nicht das angebrachte Maß an Expertise und Wissen zu Geschlechtervielfalt und Intergeschlechtlichkeit mitbrachten, um alle adäquaten Begrifflichkeiten zu kennen oder ausreichend genau zu verstehen, was die Interviewpartnerne und Autorne ursprünglich intendiert hatten. Die Expertise von Personen mit themenrelevanter gelebter Erfahrung sowie das Bewusstsein für die Bandbreite von Begriffen, um diese im Themenbereich genderfairer Sprache zu beschreiben, wird nicht zuletzt deswegen benötigt, weil sprachlich nicht nur das Ent-Gendern von Wörtern, sondern auch die Begrifflichkeiten zu Geschlecht, Geschlechtervielfalt und Sexualität zunehmend aktiv verhandelt werden. Ein dafür geeigneter Prozess, so Anita Wilson, existiert in der Übersetzungsbranche bereits in einem anderen Kontext: Beim sogenannten Cultural Review wird die korrekte Übersetzung von Spezialbegriffen in ihrem jeweiligen Kontext gesondert geprüft. Dieser Prozess wird aber für genderfaire Sprache bisher noch nicht verwendet.

### 4.2 Geringes Bewusstsein für Geschlechtervielfalt und genderfaire Sprache

Die Panelisterne waren sich einig, dass noch Sensibilisierung notwendig ist, und betonten mehrmals, dass es bei genderfairer Sprache darum geht, Menschen Respekt entgegenzubringen (eine Erkenntnis, die einige Teilnehmerne auch aus dem Workshop berichteten). Während Sprachentwicklung ein fortlaufender Prozess und immanenter Bestandteil lebender Sprachen ist (z. B. werden heutzutage einige früher verbreitete Ausdrücke und Begriffe nicht mehr als akzeptabel angesehen), passiert diese in der Regel zuerst in kleineren Kreisen, bevor neue Sprachmuster Einzug in größere Bevölkerungsgruppen finden. Hier verwiesen die Diskutanterne außerdem darauf, dass im Vergleich zum Englischen (Verwendung von singular they) Strategien zur genderfairen Sprache auf Deutsch vielfältiger und komplizierter sind, was die Verbreitung genderfairer Sprache erschwert. Die Aktualität des Themas – in den vergangenen Jahren wurden in Österreich und Deutschland die gesetzlichen Grundlagen für alternative Geschlechtseinträge geschaffen [13, 20] – wurde sowohl im Workshop als auch in der Podiumsdiskussion thematisiert.

Übersetzerne und genderfaire maschinelle Übersetzung haben das Potenzial, genderfaire Sprache einer breiteren Masse zugänglich zu machen. Erstere haben dabei eine Funktion als Brückenbauerne: Nicht nur vermitteln sie zwischen Sprachen, sie können durch ihre Arbeit auch aufkommende sprachliche Entwicklungen in die Bevölkerung tragen. Einen Weg dazu könnten literarische Übersetzungen bieten: Aus dem Publikum kam die Anregung, dass in englischsprachigen Geschichten oft Nebenfiguren vorkommen, deren Geschlecht durch de Autore nicht bestimmt wurde. Diese Figuren könnten – nach Rücksprache mit de Autore – als nicht-binär übersetzt werden. Maschinelle Übersetzung kann ebenfalls eine zentrale Rolle bei der Verbreitung genderfairer Sprache spielen, da viele Inhalte im Internet automatisch übersetzt werden, oft ohne dass dies für Benutzerne erkennbar ist.

### 4.3 Abschätzungen zwischen genderinklusiver und genderneutraler Sprache

Die Panelisterne betonten, dass ein einziger Ansatz für genderfaire Sprache nicht wünschenswert sei, da eine einzelne Norm zu starr für die dynamische Entwicklung in Sprache, Identifikation und Wahrnehmung sei. Im Workshop wurde dies mehrfach mit der Phrase „One size fails all" (im Gegensatz zu „One size fits all") zum Ausdruck gebracht, welche auch hier wieder zum Vorschein trat.

---

[19] In diesem Abschnitt verwenden wir das De-e-System des Vereins für geschlechtsneutrales Deutsch. Dabei wird im Singular grundsätzlich die Endung *-e* und im Plural die Endung *-erne* genutzt. Dementsprechend werden auch Artikel dekliniert (siehe https://geschlechtsneutral.net/dey-e-system/, abgerufen am 13. 4. 2022).

Die Wahl der Strategie, so die Experterne, sei kontextabhängig: Das Hinweisen auf Geschlecht kann in einigen Kontexten notwendig und in anderen (z. B. in Formularen oder in der Anrede unbekannter Personen) unerheblich sein. Wichtig ist auch die Möglichkeit, mit Sprache zu experimentieren, insbesondere da im Deutschen die Ansätze für genderfaires Formulieren komplexer als in Sprachen wie Englisch oder Schwedisch sind.

Eine Bemerkung im Rahmen der Podiumsdiskussion bezog sich auf den Wissensstand der Text-Rezipienterne: Nicht alle Leserne kennen jede Strategie, und das Bewusstsein für das Thema fehlt oft noch. Das Zielpublikum sollte daher nicht überfordert werden, und die Wahl der genderfairen Strategie sollte sich an dessen Wissensstand orientieren.

### 4.4 Zusammenarbeit zwischen den im Workshop involvierten Expertisen

Alle Panelisterne sahen gegenseitiges Verständnis für die Schwierigkeiten und Probleme der anderen als erforderlich für die Entwicklung einer adäquaten Lösung für genderfaire Sprache in der maschinellen Übersetzung. Insbesondere war allen bewusst, dass sich technisch eine Vielzahl an Lösungen anbietet, diese jedoch mit Aufwand verbunden sind. Dieser Aufwand sollte aber auch nicht als Ausrede verwendet werden, gar nichts zu unternehmen. Die Vertreterne der queeren Community erklärten, sich oft als Bittstellerne zu fühlen, und wünschten sich, öfter und besser beachtet und eingebunden zu werden.

Es wurde auch die Idee eines „Fair Minimums" aufgebracht: eine angemessene Behandlung und Berücksichtigung queerer und nicht-binärer Personen(gruppen) in Texten, im Gegensatz zu einem „Bare Minimum", das beispielsweise dem entspräche, „nur" nicht zu misgendern. Dieses würde ein Mindestniveau an genderfairer Sprache beschreiben, das es in einem Text zu erreichen gilt. Damit, so die Übersetzerne, könnte es leichter sein, ein breiteres Publikum zu erreichen und auch für das Thema zu begeistern oder davon zu überzeugen.

### 4.5 Genderfaire Sprachtechnologien

Genderfaire Sprachtechnologien sind bisher kaum erforscht. Tristan Miller, ein Expertere aus dem Bereich Machine Translation, erklärte, dass Gender Bias ein bekanntes Problem in Sprachtechnologien darstelle (siehe auch Abschnitt 2.2) – mehrere Beispiele von bekannten Biases kamen sowohl von Panelisterne als auch aus dem Publikum. Jedoch fehlen die Lösungen zu diesem Problem, unter anderem, weil zu wenige Daten für die Entwicklung einer genderfairen maschinellen Übersetzung verfügbar sind. Demnach fehle es also grundsätzlich an Datenmaterial, von dem aus maschinelle Ansätze zur Übersetzung arbeiten könnten.

Bias in Sprachtechnologien wurde in diesem Zusammenhang maßgeblich als ein sozio-technisches Problem beschrieben. Das bedeutet, dass hier soziale Aspekte und Implikationen ebenso beachtet werden müssen wie technische Voraussetzungen. In Gesellschaften, in denen legal und sozial der Verweis auf nicht-binäre Geschlechter notwendig wird, ist also auch ein genderfairer Sprachgebrauch in der maschinellen Übersetzung notwendig. Dementsprechend gilt es, ausreichend Datenmaterial mit unterschiedlichen genderfairen Strategien zu erstellen, um zu vermeiden, dass maschinell ein unerwünschter geschlechtsbinärer und damit exklusiver Sprachgebrauch wiederholt reproduziert wird.

Unter den Panelisterne herrschte dabei Konsens darüber, dass Sprache generell gesellschaftliche Realitäten abbildet und wiederum beeinflusst, also auch das Potenzial birgt, die Voraussetzungen für bessere Gleichstellung und Inklusion zu schaffen. Aus diesem Grund sollte aus ihrer Sicht bewusst Einfluss auf die verwendete Sprache in der Alltags- und professionellen Nutzung genommen werden. Hervorgehoben wurde auch ein rechtlicher Aspekt: Alle Personen müssen richtig erfasst werden können und die gleichen Rechte und Ansprüche auf respektvolle Kommunikation haben.

### 4.6 Genderfaire Sprache und Zugänglichkeit

Aus dem Publikum kam weiters die Anmerkung, dass Gender-Gap-Strategien (wie Gendersternchen, Unterstrich oder Doppelpunkt) das Lesen mit Screenreader-Technologien beeinträchtigen könne. Dieses Problem im Rahmen der Zugänglichkeit wurde von den Teilnehmerne im Workshop genauso wie in der Podiumsdiskussion anerkannt – notwendig sind hierbei das Berücksichtigen solcher Strategien in Screenreadern, Studien zum Thema Accessibility sowie Verständlichkeit und die Betrachtung dieser Aspekte bei der Wahl der Strategie für genderfaire Sprache. Allerdings stellte sich hier im Nachgang aufgrund einer kürzlich erschienen Studie der deutschen Überwachungsstelle des Bundes für Barrierefreiheit von Informationstechnik heraus, dass diese Strategien weniger tatsächliche Probleme verursachen als bisher angenommen [28].

Insgesamt konnten Themen des Workshops durch die Diskussion einer breiteren Öffentlichkeit vorgestellt sowie tiefergehend reflektiert werden. Dahingehend können wir aus unserer Perspektive die Planung einer derartigen Veranstaltung zum Abschluss eines solchen partizipativen Workshops sehr empfehlen, nicht zuletzt weil dies auch den Teilnehmerne noch einmal das Selbstvertrauen gab, das erworbene Wissen auf dieser Ebene zu vermitteln.

## 5 CONCLUSIO

Um die Sachlage hinsichtlich genderfairer Sprache in maschinellen Übersetzungstechnologien ausführlich zu untersuchen sowie erste adäquate Lösungsansätze zu entwickeln, führten wir einen partizipativen Workshop mit Teilnehmxs[20] aus nicht-binären und queeren Gruppen, aus der Übersetzungspraxis sowie mit Expertise in maschineller Übersetzung. Über drei Tage hinweg konnten durch diese partizipative Vorgangsweise Aspekte hinsichtlich der *Problemstellung*, der *Zielvorgaben* sowie der möglichen *Lösungsansätze* erarbeitet werden.

Im Bezug auf die Problemstellung konnten wir feststellen, dass diese von einer Mischung aus fehlendem Verständnis für genderfaire Sprachverwendung in der Gesamtgesellschaft, daraus folgender fehlender Praxis für Übersetzxs sowie fehlender Datengrundlage für maschinelles Lernen geprägt ist. Der Status quo führt allerdings zum wiederholten Ansprechen durch ein falsches Geschlecht für nicht-binäre Individuen, während die Vielfalt an Optionen wiederum zu Unsicherheit bezüglich der korrekten sprachlichen Zuschreibung bei dem nicht betroffenen Personenkreis zugehörigen Menschen führt.

---

[20] In diesem Abschnitt verwenden wir die Endung *-x*, wie sie von Hornscheidt vorgeschlagen wurde [22]. Dabei wird das „x" jeweils als [ɪks] und im Plural „xs" als [ɪksəs] ausgesprochen [1].

| Kategorie | Empfehlungen |
|---|---|
| regulatorisch | Richtlinien für genderfaire Sprache entwickeln |
| | unterschiedliche Register für genderfaire Sprache dokumentieren |
| methodisch | intrasprachliche Übersetzungen durchführen |
| | Pre- und Post-Editing Prozesse untersuchen |
| | einzelne Strategien systematisch analysieren |
| organisatorisch | Plattform für institutionalisierten Austausch zwischen Expertxgruppen schaffen |
| | Bildungsmaterialien zu genderfairer Sprache entwickeln |
| | transnationales Netzwerk aufbauen |
| technologisch | Plugin zur Analyse genderfairer Sprache anbieten |
| | intrasprachliches Übersetzungstool zum Ausprobieren unterschiedlicher Strategien entwickeln |
| | mit personalisierten genderfairen Strategien in automatisierter Übersetzung experimentieren |

**Tabelle 1: Empfehlungen zu weiteren (Forschungs-)Vorhaben im Bereich genderfairer Sprache**

Bezüglich der Zielvorgaben stellte sich heraus, dass nicht-binäre und queere Individuen eher bevorzugen, wenn vergeschlechtlichte Sprachreferenzen individuelle Präferenzen respektieren und befolgen, dass sie diese aber auch in vielerlei Hinsicht oftmals für unnötig und vermeidbar erachten. Hier stellte sich allerdings als möglicher Zielkonflikt heraus, dass die Wünsche von maschinellen wie menschlichen Übersetzxs eher in Richtung eines Standards bzw. einer Norm tendieren, was im Kontrast zu der von queeren und nicht-binären Personen gewünschten Pluralität und Offenheit für Veränderungen steht.

Hinsichtlich möglicher Lösungen wurde ein eigenständiges Berufsbild dex genderfairen Pre- und Post-Editx entwickelt und die Notwendigkeit von intralingualen Übersetzungen von nicht genderfairer Sprache in unterschiedliche Grade von genderfairer Sprache etabliert. Zudem wurden erste Überlegungen zur Korpuserstellung durchgeführt und letztlich festgehalten, dass es für marginalisierte Geschlechter in diesem Kontext konkrete Solidarität und Schulterschlüsse (auch im Hinblick auf Zugänglichkeit) benötigt. Dabei sollten personalisierbare und adaptierbare Lösungsmöglichkeiten bevorzugt werden, die auf individuelle (und sich potenziell ändernde) Präferenzen eingehen können. Wir reflektieren nun kurz über das generelle Vorgehen in dieser Forschungsarbeit und zeigen mögliche zukünftige Schritte zur Entwicklung genderfairer maschineller Übersetzungstechnologien auf.

## 5.1 Methodische Reflexionen und Einschränkungen

Trotz unserer Bemühungen im Rahmen der Einladung und Organisation des Workshops haben nur wenige Vertretxs der queeren und nicht-binären Gemeinschaften teilgenommen. Ob das an mangelndem Vertrauen in die Organisatxs, pragmatischen (z. B. ökonomischen) Faktoren oder Angst vor Tokenisierung lag, wissen wir nicht; wir vermuten aber, dass längere Vorlaufzeit und mehr Vorbereitungsarbeit sinnvoll gewesen wäre, um mehr Vertrauen in der Community zu gewinnen (vgl. z. B. [15]).

Hinsichtlich der verwendeten Methoden fiel vor allem von Seiten der „professionellen" Vertretxs aus MT- und Übersetzungskontexten eine anfängliche leichte Hemmung gegenüber der Aufforderung zum kreativen Arbeiten im Rahmen des Utopienstormings (siehe Abschnitt 3.2.3) auf, die sich aber schnell löste. Im Rahmen der Einheit zur genderfairen „Übersetzung" der Beschreibungen fiktiver Charaktere (Abschnitt 3.2.4) wurde die Aufgabenstellung von den Gruppen teils sehr verschieden interpretiert, was zu kurzer Verwirrung und abwechslungsreichen Ausarbeitungen führte. Außerdem kam es vereinzelt zu Unklarheiten bezüglich des zeitlichen Ablaufs des Workshops.

Obwohl die Stimmung generell von allen Beteiligten als wertschätzend wahrgenommen wurde, mussten (mehrfach) marginalisierte Personen teilweise wiederholt erklären, verteidigen und rechtfertigen, warum ihnen auch sprachlich Respekt entgegen gebracht werden sollte. Da viele marginalisierte Personen derartige Erfahrungen täglich machen [37], ist es demnach auch durchaus möglich, dass manche von der Teilnahme Abstand genommen haben, um genau diese unfreiwillige emotionale Arbeit bzw. das Risiko, diese durchführen zu müssen, zu vermeiden. Allys (den Gemeinschaften nahe stehende, aber nicht zugehörige, Personen mit Verständnis für deren Anliegen und Diskriminierungserfahrungen) versuchten zwar teilweise, den queeren und nicht-binären Personen diese Arbeit abzunehmen. Dennoch konnte aber aus deren Sicht bei gemischten Gruppen nicht a priori davon ausgegangen werden, dass das gegenseitige Verständnis von allen Beteiligten gegeben sein würde – was ja bis zu einem gewissen Grad gerade auch die Prämisse für unsere Arbeit darstellte. Dahingehend halten wir es für notwendig, dass den jeweiligen Expertxs dezidiert auch geschützte Räume außerhalb derart breit angelegter Kooperationen eröffnet werden, um ihre Beiträge jenseits von permanenten Aufklärungsbemühungen wissenschaftlich aufarbeiten zu können.

## 5.2 Weitere (Forschungs-)Aktivitäten

Im Verlauf des Workshops und der Podiumsdiskussion entwickelten die Teilnehmxs einigen Ideen für weiterführende Aktivitäten in Forschung und Entwicklung sowie hinsichtlich Öffentlichkeits- und Vernetzungsarbeit. Diese haben wir auch in Tabelle 1 zusammengefasst.

Technisch interessant wären die Entwicklung (1) eines öffentlich verfügbaren Tools, das die Konvertierung von längeren Textbeispielen mit allen Facetten der deutschen Sprache zwischen Strategien ermöglicht und (2) eines Browser-Plugins oder ähnlichen Tools mit genderfairer Rechtschreibprüfung, welche genderfaire Formulierungen vorschlägt, sowie (3) die Erarbeitung einer technischen Lösung für ein personalisierbares maschinelles Übersetzungssystem. Projekte wie z. B. *Gendy*[21] liefern bereits erste Schritte, hängen aber noch vollständig vom individuellen Willen einzelner Webseiten-Betreibxs ab, ein derartiges Plugin für genderfaire Sprache von sich aus aktiv einzubauen. Als Öffentlichkeits- und Vernetzungsthemen wurde während des Workshops angeregt, (4) als Bildungsmaterial für Schulen ein Kinderbuch mit Bastelanleitungen für das eierlegende Wollmilch-Wir-sind-Wir und das Lego-Einhorn Leonda, the Gender Avenger als visuelle Repräsentationen von geschlechtlicher und sprachlicher Vielfalt inklusive altersgerechter Information zu Geschlechtervielfalt und Sprache zu publizieren, (5) konkrete Plattformen für den weiteren Austausch innerhalb der drei Personengruppen des Workshops, aber auch mit interessierten anderen Personen, einzurichten, sowie (6) ein transnationales Netzwerk im deutschen Sprachraum zu organisieren. Im Gebiet der Linguistik und Übersetzung schlagen wir vor, (7) weiterführende Forschungsaktivitäten für die Überprüfung der Lesbarkeit, Verständlichkeit, Akzeptanz und Niederschwelligkeit konkreter Strategien mit verschiedenen Personenkreisen durchzuführen und (8) weiterführende Aktivitäten für die Überprüfung der Übersetzbarkeit einzelner Strategien zu setzen. Abschließend geht als Ergebnis des Workshops deutlich hervor, dass inter- und transdisziplinäre Expertxn gemeinsam an der Fragestellung von genderfairer maschineller Übersetzung arbeiten müssen, um effektiv adäquate Lösungen erarbeiten zu können, die sowohl Akzeptanz in den betroffenen Gemeinschaften finden als auch aus Anwendxsicht praktisch gut nutzbar sind.

## ACKNOWLEDGMENTS

Dieses Projekt wurde finanziell durch das Center for Technology and Society (https://cts.wien) gefördert. Wir möchten uns ganz herzlich bei unseren Mitorganisator_innen von der FH Campus Wien, Katharina Bühn, Daniela Duh, Arthur Mettinger, Igor Miladinovic und Sigrid Schefer-Wenzl, sowie bei den Teilnehmer_innen des Workshops und der Podiumsdiskussion für ihr Engagement und ihre wertvollen Ansichten und Einblicke zum Thema der genderfairen Sprache und Sprachtechnologie bedanken. Weiters herzlichen Dank an Andreas Czák für die professionelle Aufnahme und Nachbearbeitung des Videos unserer Podiumsdiskussion, an die Erste Group für die Bereitstellung der Räumlichkeiten für die Podiumsdiskussion sowie an die anonymen Reviewer:innen für viele hilfreiche Hinweise zur Verbesserung von Inhalt und Darstellung dieses Artikels.

---

[21] https://github.com/soeren-kirchner/gendy, abgerufen am 13. 4. 2022.